%% file: main.tex
\documentclass[conference]{IEEEtran}

\usepackage{etoolbox}
\makeatletter
\patchcmd{\@makecaption}
  {\scshape}
  {}
  {}
  {}
\makeatletter
\patchcmd{\@makecaption}
  {\\}
  {.\ }
  {}
  {}
\makeatother
\def\tablename{Table}
\usepackage{amsmath}
\usepackage{fancyhdr}
\usepackage{microtype}
\usepackage{graphicx}
\usepackage{subfigure}
\usepackage{float}
\usepackage{booktabs} % for professional tables
\usepackage{etoolbox}
\usepackage{enumitem}
\usepackage{hyperref}
\usepackage{cite}
\usepackage{amsmath,amssymb,amsfonts}
\usepackage{algorithmic}
\usepackage{multirow}
\usepackage{textcomp}
\usepackage{xcolor}
\fancypagestyle{alim}{\fancyhf{}\fancyfoot[L]{\textbf{*} All authors contributed equally}}

\begin{document}

\title{Merged-GHCIDR: Geometrical Approach to Reduce Image Data}
% \author{\IEEEauthorblockN{Anonymous Authors}}
\author{
\IEEEauthorblockN{\hspace{1.25cm}Devvrat Joshi*}
\IEEEauthorblockA{\hspace{1.25cm}devvrat.joshi@iitgn.ac.in}
\textit{\hspace{1.25cm}IIT Gandhinagar, India}\\
\and
\IEEEauthorblockN{\hspace{1.3cm}Janvi Thakkar*}
\IEEEauthorblockA{\hspace{1.3cm}janvi.thakkar@iitgn.ac.in}
\textit{\hspace{1.3cm}IIT Gandhinagar, India}\\
\and
\IEEEauthorblockN{\hspace{1.3cm}Siddharth Soni*}
\IEEEauthorblockA{\hspace{1.3cm}siddharth.soni@iitgn.ac.in}
\textit{\hspace{1.25cm}IIT Gandhinagar, India}\\
\and
\IEEEauthorblockN{Shril Mody*}
\IEEEauthorblockA{paresh.mody@iitgn.ac.in}
\textit{IIT Gandhinagar, India}\\
\and
\IEEEauthorblockN{Rohan Patil}
\IEEEauthorblockA{rohan.patil@iitgn.ac.in}
\textit{IIT Gandhinagar, India}\\
\and
\IEEEauthorblockN{Nipun Batra}
\IEEEauthorblockA{nipun.batra@iitgn.ac.in}
\textit{IIT Gandhinagar, India}\\
}
\maketitle

\begin{abstract}

The computational resources required to train a model have been increasing since the inception of deep networks. Training neural networks on massive datasets have become a challenging and time-consuming task. So, there arises a need to reduce the dataset without compromising the accuracy. In this paper, we present novel variations of an earlier approach called reduction through homogeneous clustering for reducing dataset size. The proposed methods are based on the idea of partitioning the dataset into homogeneous clusters and selecting images that contribute significantly to the accuracy. We propose two variations: Geometrical Homogeneous Clustering for Image Data Reduction (GHCIDR) and Merged-GHCIDR upon the baseline algorithm - Reduction through Homogeneous Clustering (RHC) to achieve better accuracy and training time. The intuition behind GHCIDR involves selecting data points by cluster weights and geometrical distribution of the training set. Merged-GHCIDR involves merging clusters having the same labels using complete linkage clustering. We used three deep learning models- Fully Connected Networks (FCN), VGG1, and VGG16. We experimented with the two variants on four datasets- MNIST, CIFAR10, Fashion-MNIST, and Tiny-Imagenet. Merged-GHCIDR with the same percentage reduction as RHC showed an increase of 2.8\%, 8.9\%, 7.6\% and 3.5\% accuracy on MNIST, Fashion-MNIST, CIFAR10, and Tiny-Imagenet, respectively.
\end{abstract}
\thispagestyle{alim}

\section{Introduction}
\input{introduction}

\section{Background}
\input{background}
% \subsection{Maintaining the Integrity of the Specifications}

% The IEEEtran class file is used to format your paper and style the text. All margins, 
% column widths, line spaces, and text fonts are prescribed; please do not 
% alter them. You may note peculiarities. For example, the head margin
% measures proportionately more than is customary. This measurement 
% and others are deliberate, using specifications that anticipate your paper 
% as one part of the entire proceedings, and not as an independent document. 
% Please do not revise any of the current designations.

\section{Preliminaries}
\input{preliminaries}

\section{Proposed Approach}
\input{proposed_approach}

\section{Evaluation}
\input{evaluation}

\section{Conclusion}
To overcome the problem of the non-parametric nature of RHC and the non-interpretability of the reduced dataset formed by the existing algorithms, we proposed two novel variations for data reduction. The novelty in our GHCIDR approach includes sampling diverse images from the entire volume of the cluster using the concept of annuli division. We further proposed Merged-GHCIDR over GHCIDR by merging the homogeneous clusters having the same labels using complete linkage clustering. Merged-GHCIDR gives an edge over existing algorithms by providing the user control over the size of the reduced dataset. Our approach outperforms RHC and the previously proposed approaches in terms of accuracy and training time on neural network models with a similar reduction rate. As validated from the experiments, GHCIDR gave accuracy similar to that of the full dataset by minimizing the tradeoff between accuracy and reduction rate. Thus, we claim that the reduced datasets formed by our approaches can replace the original datasets without compromising the overall characteristics of the datasets.

\section{Future Work}
In the future, we plan to reduce the Imagenet~\cite{deng2009imagenet} and similar large datasets using the proposed approaches. We would also compare this work with non-clustering based algorithms like Light-weight coresets~\cite{bachem2018scalable} and other sampling algorithms. Moreover, we hope to improve the algorithms by sampling the reduced dataset for robust training of neural networks against noisy labels. 

The use of the neural network in implementing real-life applications has increased exponentially in the past couple of decades, and privacy has been a major concern. While open sourcing a dataset, it is important to make it secure from privacy attacks. Thus to ensure the privacy of the reduced dataset, we aim to integrate the Merged-GHCIDR approach with differentially private mechanisms and make the dataset private.

\bibliographystyle{IEEEtran}
\bibliography{main}
% \begin{thebibliography}{00}
% \bibitem{b1} G. Eason, B. Noble, and I. N. Sneddon, ``On certain integrals of Lipschitz-Hankel type involving products of Bessel functions,'' Phil. Trans. Roy. Soc. London, vol. A247, pp. 529--551, April 1955.
% \bibitem{b2} J. Clerk Maxwell, A Treatise on Electricity and Magnetism, 3rd ed., vol. 2. Oxford: Clarendon, 1892, pp.68--73.
% \bibitem{b3} I. S. Jacobs and C. P. Bean, ``Fine particles, thin films and exchange anisotropy,'' in Magnetism, vol. III, G. T. Rado and H. Suhl, Eds. New York: Academic, 1963, pp. 271--350.
% \bibitem{b4} K. Elissa, ``Title of paper if known,'' unpublished.
% \bibitem{b5} R. Nicole, ``Title of paper with only first word capitalized,'' J. Name Stand. Abbrev., in press.
% \bibitem{b6} Y. Yorozu, M. Hirano, K. Oka, and Y. Tagawa, ``Electron spectroscopy studies on magneto-optical media and plastic substrate interface,'' IEEE Transl. J. Magn. Japan, vol. 2, pp. 740--741, August 1987 [Digests 9th Annual Conf. Magnetics Japan, p. 301, 1982].
% \bibitem{b7} M. Young, The Technical Writer's Handbook. Mill Valley, CA: University Science, 1989.
% \end{thebibliography}
% \vspace{12pt}
% \color{red}
% IEEE conference templates contain guidance text for composing and formatting conference papers. Please ensure that all template text is removed from your conference paper prior to submission to the conference. Failure to remove the template text from your paper may result in your paper not being published.

\end{document}

%% file: introduction.tex
Deep neural networks are increasingly used in everyday life, particularly speech and image recognition. Researchers worldwide have modeled more sophisticated architectural designs to tackle the Imagenet-Challenge, consequently increasing the number of parameters involved in training. This has led to an increase in the requirement of computational resources and an exponential increase in the time required to train the model on the full dataset. All the images in the dataset do not contribute sufficiently to the accuracy. We can thus separate a sample of images from the training dataset, which carries almost all the dataset's properties. This sampling will allow us to train a CNN efficiently in time and memory with maintaining the same accuracy. This paper attempts to reduce the dataset size by sampling a fraction of the whole dataset such that when trained on this reduced data, the accuracy is equivalent to that of the original dataset.

% Deep neural networks are becoming increasingly used in everyday life, particularly in speech and image recognition. Researchers from all over the world have modelled more sophisticated architectural designs in order to tackle the Imagenet Challenge, consequently increasing the number of parameters involved in training. This has led to increase in requirement of computational resources as well as exponential increase in time required to train the model on full dataset. In this paper, we attempt to reduce the dataset size by sampling a fraction of the whole dataset, so that when trained on this reduced data, the accuracy is equivalent to that of the original network.

% Training a Convolutional Neural Network (CNN) on an image dataset generally consumes a significant amount of time (in hours and days) despite the size of the dataset used. Image datasets such as Imagenet and CIFAR10 take a good amount of time to train CNNs and require ample memory space for processing. The accuracy achieved by training a CNN on a full dataset can also be achieved by training it on a subset of images from the same dataset. This is because all the images do not contribute sufficiently to the accuracy. We can separate a sample of images from the training dataset, which carries almost all the dataset's properties. This sampling will allow us to train a CNN efficiently in time and memory with maintaining the same accuracy. Our work tries to sample this subset of a dataset, simultaneously conserving the accuracy. 

Various approaches for dataset reduction have been proposed in the past, such as Reduction through homogeneous clustering (RHC)~\cite{ougiaroglou2012efficient} and Principle Sample Analysis (PSA)~\cite{inproceedings}. Our algorithm was designed using RHC as a baseline. RHC clusters the dataset by considering the labels of each data point. The dataset is partitioned into homogeneous clusters; \emph{i.e., all the data points belonging to that cluster will have the same label}. The final reduced dataset is constructed by taking only the centroids of these homogeneous clusters. The pitfall of RHC is that it disregards the importance of other images which form the cluster boundary, but boundary points play a significant role in distinguishing the labels of neighboring clusters~\cite{olvera2010new}. Larger clusters contain more images, so they should contribute more to the reduced dataset, but RHC also does not consider the varying size of clusters. 

Our work attempts to further improve the idea of RHC by imbibing the geometrical features of the clusters and introducing cluster weights to ensure the varying contribution from each cluster. We assured that the reduced dataset is a direct subset of the original dataset to ensure the interpretability of images, unlike RHC, which uses aggregation of the images in the reduced dataset. We propose two variations: Geometrical Homogeneous Clustering for Image Data Reduction (GHCIDR) [an extension to our previous work \cite{mody2022geometrical}] \& Merged-GHCIDR. GHCIDR gives weights to each homogeneous cluster according to its size and divides each cluster into annular regions. Then it samples images from the interior region of two concentric annuli. Thus, we can get a smaller representation of the entire cluster by selecting images spread over the cluster. 

The reduction rate of GHCIDR was found to be lesser than RHC. To regulate this reduction rate, we proposed Merged-GHCIDR. Merged-GHCIDR merges the homogeneous clusters of the same class using complete linkage clustering.
These two variants were tested on four image datasets: MNIST~\cite{deng2012mnist}, CIFAR10 ~\cite{krizhevsky2009learning}, Fashion-MNIST~\cite{xiao2017fashion} and Tiny-Imagenet~\cite{deng2009imagenet}. We tested our reduced datasets on three deep learning models; namely, Fully Connected Networks FCN~\cite{long2015fully}, VGG1~\cite{simonyan2014very} and VGG16 ~\cite{simonyan2014very}. We found both variants performed better than the baseline RHC in terms of accuracy. Merged-GHCIDR gave an accuracy of 99.04\%, 86.33\%, 72.07\% \& 55.55\% accuracy on MNIST, Fashion-MNIST, CIFAR10 and Tiny-Imagenet respectively trained on VGG model. 

%Our code  \footnote{Here is the link to the code: \textbf{\url{https://github.com/SoniSiddharth/Merged-GHCIDR}}.} is fully reproducible.
Our main contribution includes: 
\begin{itemize}
    \item We propose two novel approaches, which involve reducing the dataset size by selecting a subset of points from the original dataset. In comparison to existing works, which reduce the dataset size by modifying the data points, our work focuses on a sampling of data points based on the geometrical aspect of the training set. It outperforms the existing works, providing better accuracy with comparable reduction.
    \item Existing work on image dataset reduction uses compression to reduce the size but simultaneously compromises their interpretability, while in our approach, we sample the images that contribute significantly to the accuracy. Thus, ensuring that the images in the reduced dataset are direct subsets of the full image dataset.
   \item Unlike RHC, our algorithm is parametric in nature, i.e., the user can control the size of the reduced dataset. We used complete linkage clustering to merge the homogeneous clusters having the same labels, as we experimentally concluded that it outperforms the other agglomerative clustering techniques (Single Linkage and Average Linkage).
\end{itemize}

%% file: background.tex
In this section, we describe some of the previously proposed approaches of dataset reduction, such as Prototype Selection by Clustering (PSC), Reduction by Space Partitioning (RSP3), Reduction Through Homogeneous Clustering (RHC) and Dataset Distillation. The Reduction by Space Partitioning (RSP) algorithm \cite{sanchez2004high} is a group of approaches for reducing a dataset. The main idea of RSP1 and RSP2 is to recursively split the dataset into two sets and select the centroid of each class from the subset. RSP3, on the other hand, divides the dataset until all of the points inside a subset have the same labels. Due to the recursive nature of RSP, the size of the reduced dataset is not determined; hence, this makes it difficult for the user to get a better trade-off between accuracy and reduction. In another approach, PSC \cite{olvera2010new}, a selection algorithm uses the $k$-means clustering to extract the clusters in the original dataset. For homogeneous clusters, it selects items nearest to the centroid, while for non-homogeneous clusters, it samples the images that form the basis of separation between the different classes of the clusters. The drawback of PSC is that it requires the user to employ a trial-and-error method to determine the final count of the clusters. 

RHC improved on the weakness of PSC and RSP by providing a better trade-off between accuracy and reduction. It focuses on constructing homogeneous clusters which contain data points belonging to the same class. Initially, the algorithm considers the full training dataset as a single cluster. There are two possibilities for the clusters on each iteration - homogeneous and non-homogeneous. If the cluster is homogeneous, its centroid gets added to the reduced dataset (final output). If it is non-homogeneous, then we recursively cluster the points using $k$-means~\cite{chen2009k} algorithm until all the clusters become homogeneous or there is no cluster left. RHC was tested on various annotated datasets ~\cite{alcala2011keel} - Letter recognition (LR), Pen-Digits (PD), and Landsat Satellite (LS). It gave better accuracy with lesser time complexity than classical ML algorithms such as $k$-NN ~\cite{guo2003knn}, and PSC ~\cite{olvera2010new}.

Recently, Dataset Distillation \cite{wang2018dataset} approach proposed by Wang \textit{et al.}, an alternative formulation for the model distillation (a widely used technique in ensemble learning \cite{radosavovic2018data}), aims to refine the knowledge of a large dataset to a smaller one. The idea was to synthesize a smaller dataset (points not necessarily from the original dataset) such that when the model is trained on distilled data, it gives accuracy almost the same as that of the model trained on original data. Dataset distillation approach \cite{wang2018dataset} when applied on MNIST comprising 60000 images, reduced the dataset to only ten images, i.e., as low as one image per class. However, this dataset has a limitation that we cannot identify the class of the image, making it less interpretable.

\begin{figure}
\centering
% \subfloat
{\includegraphics[width=3.5cm]{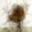}}%
\qquad
% \subfloat
{\includegraphics[width=3.5cm]{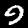}}%
\caption{Non Human-readable images of RHC (left: CIFAR10 (class: bird), right: MNIST (class: 9))}%
\label{fig:nonhuman}%
% \vspace{-2}
\end{figure}
% Training these 10 images on traditional LeNET [7] gave an accuracy of 94\%.
% Motivation is included in the paragraph below.

In this work, we took RHC as baseline and proposed our own approach which overcomes the problem of ~\cite{olvera2010new}, \cite{sanchez2004high}, \cite{wang2018dataset} and \cite{ougiaroglou2012efficient}. We applied the RHC algorithm over image datasets and discovered it performed poorly for images. Since the images in the reduced set are an aggregation of a cluster, they become blurry, and we cannot effectively determine the class of the centroid as shown in \textbf{Figure:} \ref{fig:nonhuman}. Also, the RHC algorithm is a non-parametric algorithm where the size of the reduced set is not determined, and therefore we cannot get a specific desired reduction. RHC has its own limitations, which can be improved with an enhanced approach to data point selection. This work proposes the Merged-GHCIDR, a parametric algorithm that outputs data with a precise reduction rate, forming the subset of the full dataset.

%When we tested RHC on image datasets, we found that the centroids of the images could be non human-readable as shown in \textbf{Figure:} \ref{fig:nonhuman}. 
%This aggregation of images reduced the final accuracy, and thus we modified the RHC algorithm to output human-readable images giving high accuracy.

% As noted in the introduction, the "\verb|acmart|" document class can
% be used to prepare many different kinds of documentation --- a
% double-blind initial submission of a full-length technical paper, a
% two-page SIGGRAPH Emerging Technologies abstract, a "camera-ready"
% journal article, a SIGCHI Extended Abstract, and more --- all by
% selecting the appropriate {\itshape template style} and {\itshape
% template parameters}.
% This document will explain the major features of the document
% class. For further information, the {\itshape \LaTeX\ User’s Guide} is
% available from
% \url{https://www.acm.org/publications/proceedings-template}.

%% file: preliminaries.tex
\textbf{Definition 1 (Complete-linkage clustering)}.
\textit{This is one of the agglomerative clustering algorithms. Initially, all the points (outputs from GHCIDR) are considered single element clusters. Then iteratively, clusters are merged with the following invariant - merge clusters with the minimum distance between the farthest points of any two clusters.
Mathematically, the complete-linkage involves calculating the distance $S(A,B)$ between the clusters A and B as the following expression: 
\begin{equation*}
    S(A,B) = \max_{{a\in A, b\in B}} d(a,b)
\end{equation*}
where, $d(a,b)$ is the distance between the elements $a\in A, b\in B$ and A,B are the clusters.
}

\textbf{Definition 2 (N-dimensional ball) \cite{enwiki:1065144473}}.
\textit{A n-sphere of radius r is defined as a set of points in (n + 1)-dimensional Euclidean space that is at a constant distance r from a fixed point c, where r can be any positive real integer and c can be any point in (n + 1)-dimensional space for any natural number n.
\begin{equation*}
    S^{n}(r) = \{x \in \mathbb{R}^{n+1}: \vert\vert x \vert\vert =r\}
\end{equation*}
}
\textbf{Definition 3 (Annulus)}.
\textit{The volume between two concentric n-spherical surfaces is defined as the annulus. In this paper, we divide the n-sphere into multiple concentric annuli and use this representative structure to sample the points based on their geometrical distribution inside the annulus.}

%% file: proposed_approach.tex
In this section, we propose the approach to reduce the dataset by sampling the data points that contribute significantly to the accuracy. We further build upon the central idea of the RHC algorithm that clusters the points based on the homogeneous labels and generates a reduced dataset by taking the centroid of each cluster. The images in this reduced dataset consist of centroids, which are an aggregation of all the points representing the cluster, making the images unidentifiable. RHC ignores boundary points of each cluster which act as a distinguishing factor for classification between images of two adjacent clusters ~\cite{olvera2010new}. Also, RHC avoids the size of the clusters making the contribution of different-sized clusters equal in the reduced dataset.

To overcome these problems, in this work, we propose - GHCIDR and Merged-GHCIDR. Geometrical Homogeneous Clustering for Image Data Reduction (GHCIDR) creates homogeneous clusters similar to the concept of RHC. However, here we imbibe the geometrical aspect of the dataset while RHC considers only the centroid. GHCIDR takes into account the size of clusters by assigning weights to decide the number of data points to be selected from each cluster. Hence, the images in the reduced dataset of GHCIDR are identifiable and a better representation of the original dataset. Another drawback of RHC is that it is non-parametric, and the size of the reduced dataset is not in the user's control. We further improved the GHCIDR algorithm into Merged-GHCIDR. In this approach, we merged the clusters obtained from GHCIDR using complete linkage clustering to give users control over the size of the reduced data, making the algorithm parametric. This overcomes the problem encountered in RHC and GHCIDR.

% his algorithm is parametric in nature thus, the size of the reduced dataset can be altered as needed.
% Another drawback of RHC is that it is non-parametric and the size of reduced dataset is not in the control of the user. 

% The images in the reduced dataset consists of centroids which are an aggregation of all the points representing the cluster, making (images)them unidentifiable.

% This aggregation of images is blurry . RHC ignores the size of clusters and selects the centroid of each cluster. This makes the number of images taken from small and large clusters equal in the reduced dataset. Ideally, images selected from each cluster should be proportional to the size of the cluster. Because large-sized clusters require more images to represent them and vice-versa.
% contain more images, so more images are required to represent them and vice versa.
% RHC ignores boundary points of each cluster which act as a distinguishing factor for classification between images of two adjacent clusters ~\cite{olvera2010new}. These drawbacks are caused by selecting centroids of homogeneous clusters. Thus, we propose two variants that overcome these limitations.
% \begin{figure}
% \centering
% % \subfloat
% {\includegraphics[width=3.5cm]{average.jpg}}%
% \qquad
% % \subfloat
% {\includegraphics[width=3.5cm]{imgMNIST.jpg}}%
% \caption{Non Human-readable images of RHC (left: CIFAR10 (class: bird), right: MNIST (class: 9))}%
% \label{fig:nonhuman}%
% % \vspace{-2}
% \end{figure}

\begin{figure}
\includegraphics[width=\linewidth , height=2in]{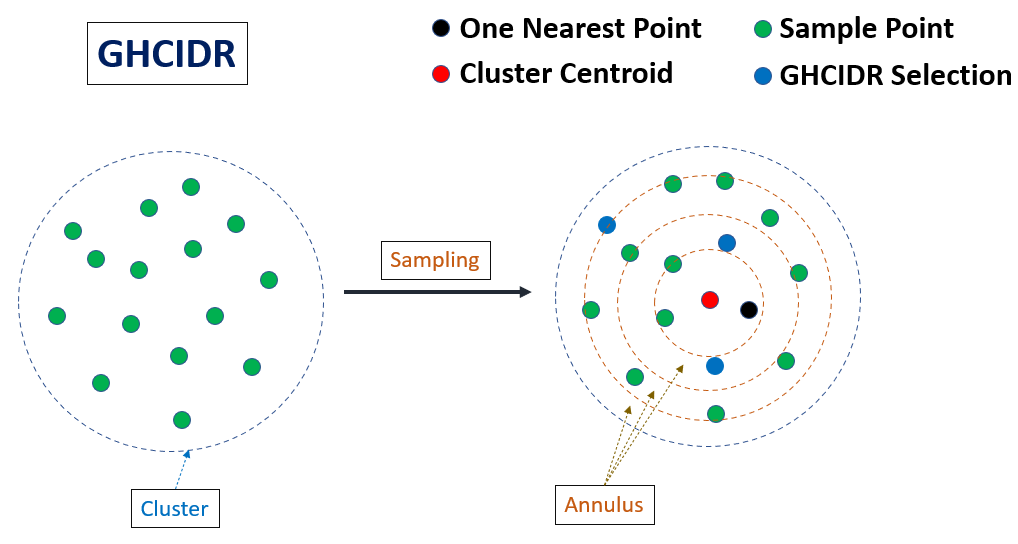}
\caption{GHCIDR: This is one of the homogeneous clusters. Divide the cluster into different annular regions and select the points nearest to the average distance of the corresponding annulus. The intuition is to select points from the entire volume of the cluster.}
% \Description{Annulus}
\label{fig:annulus.PNG}
\end{figure}

\subsection{\textbf{Geometrical Homogeneous Clustering for Image Data Reduction (GHCIDR)}}
In GHCIDR (shown in \textbf{Figure:} \ref{fig:annulus.PNG}), we introduce the idea of cluster weights by assigning more weightage to large-sized clusters. Adding cluster weights ensures we select images proportional to the cluster's size. Furthermore, RHC selects only the centroid from each homogeneous cluster, whereas GHCIDR focuses on selecting images from the entire homogeneous cluster. To incorporate RHC's concept of \textbf{centrality} in our algorithm, we select an image nearest to the cluster's centroid. In addition, we consider the cluster as an "N-dimensional ball." We divide this ball into annular regions and select images from each "annulus." This ensures that we select images from the entire volume of the cluster.

\textbf{Algorithm:} Consider the cluster as an $N$-dimensional ball and images as points in this ball.
\begin{itemize}[noitemsep,topsep=0pt]
\item Add the nearest point from the centroid of the cluster into the reduced set.
\item Let $maxDist$ be the distance of the farthest point from the centroid.
\item Let $\alpha$ be the parametric reduction rate input to the algorithm.
\item Let $size \,of \,cluster$ be the number of data points in that cluster.
\item Let $\gamma$ depend on the size of the cluster, $maxDist$ and the reduction rate $\alpha$.
$$\gamma = \frac{maxDist}{((1-\alpha)\times(size \,of \,cluster))}$$
\item Divide $maxDist$ into partitions of equal size to get $(maxDist/\gamma)$ annuli of the ball, with the innermost part being a sphere.
\item For each annulus $i$, let the inner radius and outer radius of the annulus be $R1_i$ and $R2_i$ respectively.
\item Find the point belonging to annulus $i$ having distance from the centroid nearest to $(R1_i + R2_i)/2$ and add it to reduced set.
\end{itemize}

\begin{figure}
\includegraphics[width=\linewidth]{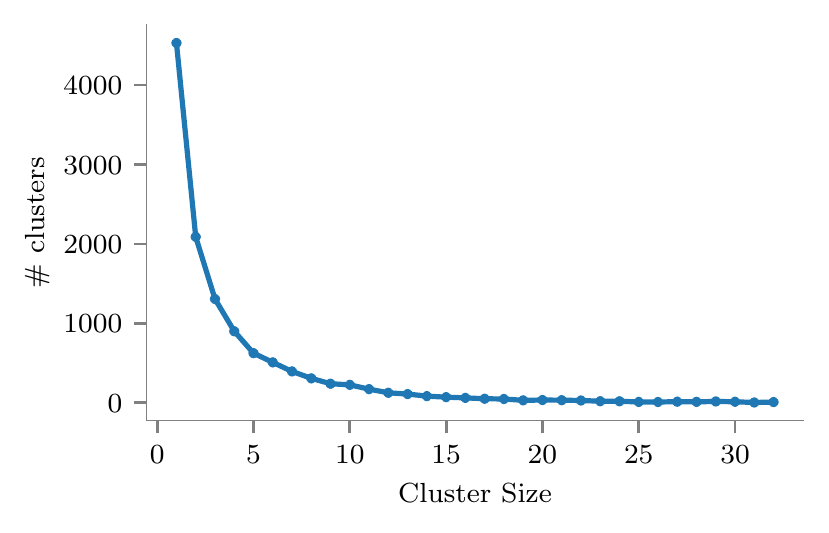}
\caption{No. of Clusters vs Cluster Size. Clusters with small sizes are large in number, and their count decreases exponentially with their size. 10000 smaller clusters contain 15000 images, whereas 2200 larger clusters contain the remaining 35000 images.}
% \Description{Cluster Size}
\label{fig:cluster.PNG}
% \vspace{-2}
\end{figure}

% \begin{figure}
% \includegraphics[width=\linewidth , height=2in]{merge.png}
% \caption{Merged-GHCIDR: Clusters having same label in vicinity are merged into one homogeneous cluster using complete linkage clustering. The intuition is to reduce the number of homogeneous clusters to achieve reduction equivalent to that of RHC.}
% % \Description{Annulus}
% \label{fig:merge.PNG}
% \end{figure}
\subsection{\textbf{Merged-GHCIDR}}
% To compare any two reduction algorithms, we need to have same reduction
GHCIDR takes the homogeneous clusters formed by the RHC as input. RHC selects only one image from each cluster, while GHCIDR selects at least one image (nearest to centroid) and more (evenly distributed across the cluster) based on the size of the cluster. Due to this, the reduction percentage of GHCIDR is always lesser than or equal to RHC. From \textbf{Figure}: \ref{fig:cluster.PNG}, we can see that as the cluster size increases, their count decreases drastically. Out of 12,200 clusters of CIFAR 10, nearly 10,000 clusters lie in a size range between 1 and 5. Because of this, the weightage assigned to small clusters becomes infinitesimal, resulting in few or zero data points being sampled from these clusters, which play a significant role in the overall representation of the dataset. Also, selecting zero images would reduce the accuracy as the number of smaller clusters is very large. To solve this tradeoff, we merge the homogeneous clusters having the same labels using complete linkage clustering ~\cite{dawyndt2005complete} (\textbf{Figure:} \ref{fig:merge.PNG}). Merged-GHCIDR applies GHCIDR on these merged clusters. We experimentally found that complete linkage clustering performed better than the other agglomerative clustering methods ~\cite{mullner2011modern} inaccuracy. 
\\
\textbf{Algorithm:}
\begin{itemize}[noitemsep,topsep=0pt]
\item Let us define $\beta$ as the ratio:
\begin{equation*}
    \begin{split}
        \beta = \frac{\#Clusters}{\#total\ Homogeneous\ Clusters\ From\ RHC}
    \end{split}
\end{equation*}
% $$$$
\item Iterate over each class and reduce the number of clusters to $(\beta \times \#total\ Homogeneous\ Clusters \ Of\ Each\ Class)$ using complete linkage clustering.
\item Apply the GHCIDR algorithm on the merged clusters to get the final condensed set.
\item Tune the value of $\beta$ such that we get the reduction rate equivalent to RHC.
\end{itemize}
\textbf{Note:} Similar reduction rate of Merged-GHCIDR and RHC helps us compare the performance of these algorithms effectively.

\input{table}

%% file: table.tex
\begin{table*}[]
\caption{\textmd{Accuracy on reduced data, Variance and \% reduction of MNIST dataset trained on FCN and VGG1 model for GHCIDR, Merged-GHCIDR and RHC.}}
\begin{center}
\begin{tabular}{|l|rrrrrrr|}
\hline
\textbf{} & \multicolumn{7}{c|}{\textbf{MNIST}} \\ \hline
\textbf{} & \multicolumn{3}{c|}{\textbf{FCN}} & \multicolumn{3}{c|}{\textbf{VGG1}} & \multicolumn{1}{c|}{\multirow{2}{*}{\textbf{\begin{tabular}[c]{@{}c@{}}\\\%Reduction\\ Rate\end{tabular}}}} \\ \cline{1-7}
\textbf{Algorithm Name} & \multicolumn{1}{l|}{\begin{tabular}[c]{@{}l@{}}\%Accuracy on \\    reduced data\end{tabular}} & \multicolumn{1}{l|}{Variance} & \multicolumn{1}{c|}{Training Time} & \multicolumn{1}{l|}{\begin{tabular}[c]{@{}l@{}}\%Accuracy on \\     reduced data\end{tabular}} & \multicolumn{1}{c|}{Variance} & \multicolumn{1}{c|}{Training Time} & \multicolumn{1}{c|}{} \\ \hline
\textbf{Full Dataset} & \multicolumn{1}{r|}{98.502} & \multicolumn{1}{r|}{0.001} & \multicolumn{1}{r|}{455s 163ms} & \multicolumn{1}{r|}{99.550} & \multicolumn{1}{r|}{0.003} & \multicolumn{1}{r|}{705s 668ms} & 0 \\ \hline
\textbf{RHC} & \multicolumn{1}{r|}{94.418} & \multicolumn{1}{r|}{0.338} & \multicolumn{1}{r|}{104s 192ms} & \multicolumn{1}{r|}{96.284} & \multicolumn{1}{r|}{0.778} & \multicolumn{1}{r|}{102s 118ms} & 95.068 \\ \hline
\textbf{Merged - GHCIDR} & \multicolumn{1}{r|}{94.574} & \multicolumn{1}{r|}{0.226} & \multicolumn{1}{r|}{106s 647ms} & \multicolumn{1}{r|}{99.040} & \multicolumn{1}{r|}{0.001} & \multicolumn{1}{r|}{101s 440ms} & 95.068 \\ \hline
\textbf{GHCIDR} & \multicolumn{1}{r|}{97.628} & \multicolumn{1}{r|}{0.021} & \multicolumn{1}{r|}{121s 351ms} & \multicolumn{1}{r|}{99.338} & \multicolumn{1}{r|}{0.002} & \multicolumn{1}{r|}{134s 325ms} & 87.273 \\ \hline
\end{tabular}
\end{center}
\label{table:MNIST}
\end{table*}

\begin{table*}[]
\caption{\textmd{Accuracy on reduced data, Variance and \% reduction of FMNIST dataset trained on FCN and VGG1 model for GHCIDR, Merged-GHCIDR and RHC.}}
\begin{center}
\begin{tabular}{|l|rrrrrrr|}
\hline
\textbf{} & \multicolumn{7}{c|}{\textbf{FMNIST}} \\ \hline
\textbf{} & \multicolumn{3}{c|}{\textbf{FCN}} & \multicolumn{3}{c|}{\textbf{VGG1}} & \multicolumn{1}{c|}{\multirow{2}{*}{\textbf{\begin{tabular}[c]{@{}c@{}}\\\%Reduction\\  Rate\end{tabular}}}} \\ \cline{1-7}
\textbf{Algorithm Name} & \multicolumn{1}{l|}{\begin{tabular}[c]{@{}l@{}}\%Accuracy on \\     reduced data\end{tabular}} & \multicolumn{1}{c|}{Variance} & \multicolumn{1}{c|}{Training Time} & \multicolumn{1}{l|}{\begin{tabular}[c]{@{}l@{}}\%Accuracy on \\    reduced data\end{tabular}} & \multicolumn{1}{c|}{Variance} & \multicolumn{1}{l|}{Training Time} & \multicolumn{1}{c|}{} \\ \hline
\textbf{Full Dataset} & \multicolumn{1}{r|}{90.034} & \multicolumn{1}{r|}{0.076} & \multicolumn{1}{r|}{472s 434ms} & \multicolumn{1}{r|}{93.338} & \multicolumn{1}{r|}{0.015} & \multicolumn{1}{r|}{704s 787ms} & 0 \\ \hline
\textbf{RHC} & \multicolumn{1}{r|}{71.786} & \multicolumn{1}{r|}{1.936} & \multicolumn{1}{r|}{143s 222ms} & \multicolumn{1}{r|}{79.102} & \multicolumn{1}{r|}{2.154} & \multicolumn{1}{r|}{104s 961ms} & 90.930 \\ \hline
\textbf{Merged - GHCIDR} & \multicolumn{1}{r|}{80.726} & \multicolumn{1}{r|}{0.619} & \multicolumn{1}{r|}{147s 197ms} & \multicolumn{1}{r|}{86.334} & \multicolumn{1}{r|}{0.784} & \multicolumn{1}{r|}{102s 131ms} & 90.995 \\ \hline
\textbf{GHCIDR} & \multicolumn{1}{r|}{87.038} & \multicolumn{1}{r|}{0.106} & \multicolumn{1}{r|}{193s 458ms} & \multicolumn{1}{r|}{91.564} & \multicolumn{1}{r|}{0.02} & \multicolumn{1}{r|}{148s 523ms} & 76.808 \\ \hline
\end{tabular}
\end{center}

\label{table:FMNIST}
\end{table*}

% \begin{table*}[]
% \begin{tabular}{|l|r|r|r|r|r|}
% \hline
%  & \multicolumn{5}{c|}{\textbf{FMNIST}} \\ \hline
%  & \multicolumn{2}{c|}{\textbf{FCN}} & \multicolumn{2}{c|}{\textbf{VGG1}} & \multicolumn{1}{c|}{\multirow{2}{*}{\%Reduction Rate}} \\ \cline{1-5}
% \textbf{Algorithm name} & \multicolumn{1}{l|}{\%Accuracy on reduced data} & \multicolumn{1}{l|}{Variance} & \multicolumn{1}{l|}{\%Accuracy on reduced data} & \multicolumn{1}{l|}{Variance} & \multicolumn{1}{c|}{} \\ \hline
% \textbf{Full Dataset} & 90.034 & 0.076 & 93.338 & 0.015 & - \\ \hline
% \textbf{RHC} & 71.786 & 1.936 & 79.102 & 2.154 & 90.930 \\ \hline
% \textbf{Merged-GHCIDR} & 80.726 & 0.619 & 86.334 & 0.784 & 90.995 \\ \hline
% \textbf{GHCIDR} & 87.038 & 0.106 & 91.564 & 0.020 & 76.808 \\ \hline
% \end{tabular}
% \caption{\textmd{Accuracy on reduced data, Variance and \% reduction of FMNIST dataset trained on FCN and VGG1 model for GHCIDR, merged-GHCIDR and RHC.}}
% \label{table:FMNIST}
% \end{table*}

\begin{table*}[]
\caption{Accuracy on reduced data, Variance and \% reduction of CIFAR10 and Tiny Imagenet datasets trained on VGG1 and VGG16 model for GHCIDR, Merged-GHCIDR and RHC respectively.}
\begin{center}
\begin{tabular}{|l|rrrr|rrrr|}
\hline
\textbf{} & \multicolumn{4}{c|}{\textbf{CIFAR}} & \multicolumn{4}{c|}{\textbf{TinyImagenet}} \\ \hline
\textbf{} & \multicolumn{4}{c|}{\textbf{VGG1}} & \multicolumn{4}{c|}{\textbf{VGG16}} \\ \hline
\textbf{Algorithm Name} & \multicolumn{1}{c|}{\begin{tabular}[c]{@{}c@{}}\%Accuracy on \\    reduced data\end{tabular}} & \multicolumn{1}{c|}{Variance} & \multicolumn{1}{c|}{\begin{tabular}[c]{@{}c@{}}\%Reduction \\  Rate\end{tabular}} & \multicolumn{1}{c|}{\begin{tabular}[c]{@{}c@{}}Training \\ Time\end{tabular}} & \multicolumn{1}{c|}{\begin{tabular}[c]{@{}c@{}}\%Accuracy on \\   reduced data\end{tabular}} & \multicolumn{1}{c|}{Variance} & \multicolumn{1}{c|}{\begin{tabular}[c]{@{}c@{}}\%Reduction \\ Rate\end{tabular}} & \multicolumn{1}{c|}{\begin{tabular}[c]{@{}c@{}}Training \\ Time\end{tabular}} \\ \hline
\textbf{Full Dataset} & \multicolumn{1}{r|}{82.658} & \multicolumn{1}{r|}{0.007} & \multicolumn{1}{r|}{0} & 601s 186ms & \multicolumn{1}{r|}{61.862} & \multicolumn{1}{r|}{0.588} & \multicolumn{1}{r|}{0} & 6550s 821ms \\ \hline
\textbf{RHC} & \multicolumn{1}{r|}{64.478} & \multicolumn{1}{r|}{2.023} & \multicolumn{1}{r|}{75.670} & 402s 921ms & \multicolumn{1}{r|}{52.066} & \multicolumn{1}{r|}{0.239} & \multicolumn{1}{r|}{47.333} & 3940s 415ms \\ \hline
\textbf{Merged - GHCIDR} & \multicolumn{1}{r|}{72.066} & \multicolumn{1}{r|}{0.208} & \multicolumn{1}{r|}{75.668} & 401s 238ms & \multicolumn{1}{r|}{55.554} & \multicolumn{1}{r|}{0.289} & \multicolumn{1}{r|}{47.280} & 3988s 761ms \\ \hline
\textbf{GHCIDR} & \multicolumn{1}{r|}{80.646} & \multicolumn{1}{r|}{0.104} & \multicolumn{1}{r|}{32.244} & 560s 150ms & \multicolumn{1}{r|}{60.492} & \multicolumn{1}{r|}{0.734} & \multicolumn{1}{r|}{15.548} & 6075s 516ms \\ \hline
\end{tabular}
\end{center}
\label{table:CIFAR10_imagenet}
\end{table*}

%% file: evaluation.tex
\subsection{Datasets}

\begin{figure}
\includegraphics[width=\linewidth , height=2in]{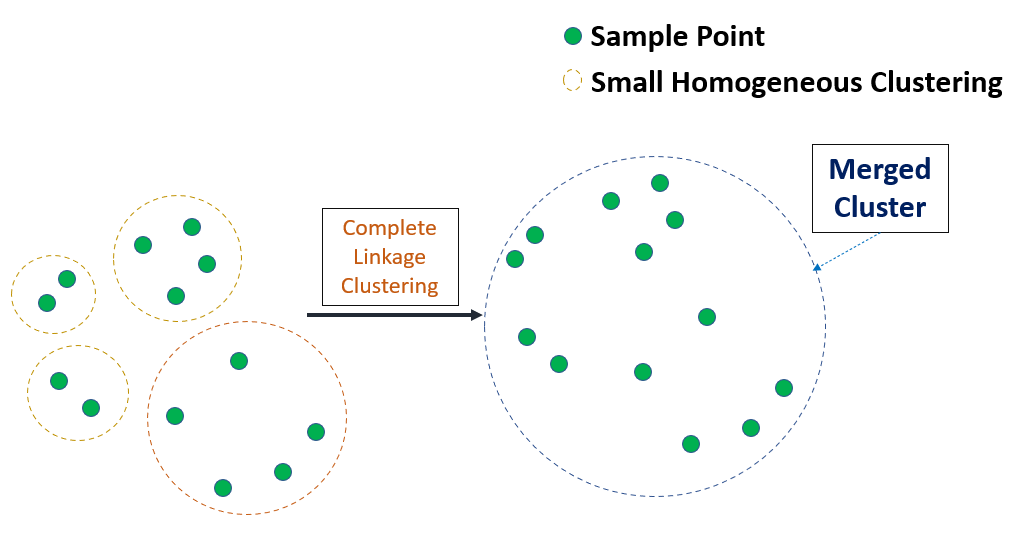}
\caption{Merged-GHCIDR: Clusters having the same label in the vicinity are merged into one homogeneous cluster using complete linkage clustering. The intuition is to reduce the number of homogeneous clusters to achieve a reduction equivalent to RHC.}
% \Description{Annulus}
\label{fig:merge.PNG}
\end{figure}

\textbf{MNIST Dataset \cite{deng2012mnist}:} It is a collection of 60,000 handwritten images with 10 classes. Dimension of each image is 28x28 pixels.

\textbf{FMNIST Dataset \cite{xiao2017fashion}:} It is a collection of 60,000 clothes images with 10 classes. Dimension of each image is 28x28 pixels.

\textbf{CIFAR10 Dataset \cite{krizhevsky2009learning}:} It is a dataset of 50,000 coloured images with 10 classes. Dimension of each image is 32x32x3 pixels.

\textbf{Tiny-Imagenet Dataset \cite{deng2009imagenet}:} It is a subset of Imagenet consisting of 100,000 coloured images with 100 classes. Dimension of each image is 64x64x3 pixels. \footnote{You can find the data on this link: \textbf{\url{https://www.kaggle.com/c/thu-deep-learning/data}}}

\subsection{Metrics}
\textbf{Accuracy:} To compare the performance of the reduced dataset and the original dataset, we have used accuracy as the performance metric. Accuracy is defined as the ratio of the total number of correct predictions to the total number of predictions on the test dataset.

\begin{equation*}
    Accuracy\ (Test\ Dataset) = \frac{\# Correct\ Predictions}{\# Total\ Predictions}
\end{equation*}
\textbf{Training Time:} It is the time taken by the Neural Network to get trained on the given dataset. We compare the training time of the neural networks trained on both the original dataset and the reduced dataset.

% % % Distance Metrics: We used various distance metrics for evaluating the distance between 2 images like`$\ell 1$norm and $\ell 2$ norm. It was found that L2 norm performs better than L1 the norm in all experiments
\subsection{Experimental Settings}
\begin{itemize}[noitemsep,,topsep=0pt]
\item We experimented on three deep learning models, namely FCN (Fully Connected Network), VGG1 and VGG16. We used Nvidia P100 GPU with 60 GB RAM for dataset reduction and training.
\item For FCN, VGG1, and VGG16, we used a batch size of 64, 100 epochs, and a learning rate of 0.001 (VGG1) and 0.01 (VGG16).
\item Each experiment was carried out five times, and average accuracy and variance were reported.
\item To get the size of condensed dataset equal to that of RHC in merged-GHCIDR, we found the optimum value of ($\beta$, $\alpha$) to be (0.4, 0.85) for MNIST, (0.3, 0.4) for CIFAR10, (0.38, 0.9) for FMNIST, and (0.5, 0.4) for Tiny-Imagenet datasets.
% \item We sampled a \emph{random} subset of images whose size was equal to the reduced dataset. For example, in Table~ \ref{table:FCN}, RHCKON on MNIST reduces dataset size by 90.13\% or samples 9.87\% of the dataset. We select 9.87\% of the points randomly in the random sampling baseline. We report that the average accuracy on randomly sampled datasets was calculated by varying the random seed.
% \item We evaluated the testing datasets and reported the accuracy for models trained on reduced data and random data.

\end{itemize}
\subsection{Results}
%3 para -- full rhc mghcidr --
The proposed algorithms were evaluated on the four datasets, and the reduced datasets were trained on FCN, VGG1, and VGG16 models. GHCIDR selects one or more images evenly from each homogeneous cluster. Thus, its reduction rate is lesser than RHC. The results were tabulated in \textbf{Tables}
\ref{table:MNIST}, \ref{table:FMNIST} and \ref{table:CIFAR10_imagenet}. 
% GHCIDR takes the homogeneous clusters formed by the RHC as input. RHC selects only one image from each cluster, while GHCIDR selects at least one image (nearest to centroid) and more (evenly distributed across the cluster) based on the size of the cluster.
Merged-GHCIDR outperformed RHC in every experiment with the same reduction rate.
\begin{itemize}[noitemsep,topsep=0pt]
\item For FMNIST dataset trained on FCN model and VGG1 model, we got a 8.940\% and a 7.232\% increase in accuracy over RHC with Merged-GHCIDR.
\item For MNIST dataset trained on FCN model and VGG1 model, we got a 0.156\% and a 2.756\% increase in accuracy over RHC with Merged-GHCIDR.
\item For CIFAR10 dataset trained on VGG1 model, we got a 7.588\% increase in accuracy over RHC with Merged-GHCIDR.
\item For Tiny-Imagenet trained on the VGG16 model; we got a 3.488\% increase in accuracy over RHC with Merged-GHCIDR.
\end{itemize}

The reduced data generated from our algorithm takes significantly less training time in comparison to the full dataset. Training time on different models were tabulated in \textbf{Tables} \ref{table:MNIST}, \ref{table:FMNIST} and \ref{table:CIFAR10_imagenet}.
Merged-GHCIDR gave almost similar training time as that of RHC for each dataset and model.

\begin{itemize}[noitemsep,topsep=0pt]
\item For MNIST dataset trained on FCN and VGG1 model, Merged-GHCIDR reduced the training time by 76.5\% and 85.6\% respectively. 
\item For FMNIST dataset trained on FCN and VGG1 model, Merged-GHCIDR reduced the training time by 68.8\% and 78.4\% respectively.
\item For CIFAR10 dataset trained on VGG1 model, Merged-GHCIDR reduced the training time by 33.2\%. 
\item For TinyImagenet dataset trained on VGG16 model, Merged-GHCIDR reduced the training time by 39.1\%. 
\end{itemize}
\textbf{Note:} The results were calculated using the time module of Python 3.6.
% Accuracy of full MNIST dataset when trained on FCN is 98.502\%. When reduced with Merged-GHCIDR, it gave an accuracy of 97.\% and a reduction of 87.27\%.
% \item Accuracy of full Fashion-MNIST dataset of Fashion-MNIST, on FCN is 87.41\%. When reduced with GHCIDR, it gave an accuracy of 83.96\% and a reduction of 76.80\%.
% From Table \ref{table:FMNIST}, we can observe that:
% \begin{itemize}[noitemsep,,topsep=0pt]
% \item Accuracy of full dataset of MNIST on VGG1 is 99.51\%. When reduced with GHCIDR, it gave an accuracy of 99.35\% and a reduction of 87.27\%.
% \item Accuracy of full dataset of Fashion-MNIST on VGG1 is 93.25\%. When reduced with GHCIDR, it gave an accuracy of 91.66\% and a reduction of 76.80\%.
% \end{itemize}
% From Table \ref{table:CIFAR_VGG}, we can observe that:
% \begin{itemize}[noitemsep,,topsep=0pt]
% \item Accuracy of full dataset of CIFAR10 on VGG1 is 82.87\%. When reduced with GHCIDR, it gave an accuracy of 81.10\% and a reduction of 32.34\%.
% \end{itemize}
\subsection{Analysis}
% variance of RHC is greater than M-ghcidr
%
Both the proposed approaches - GHCIDR and Merged-GHCIDR outperformed the baseline RHC in terms of accuracy. Merged-GHCIDR showed a significant increase in accuracy with a similar reduction rate as that of RHC. For MNIST, FMNIST, and CIFAR10 datasets, the RHC algorithm showed more variance than Merged-GHCIDR and GHCIDR. This is because RHC uses aggregation of images in its final dataset, while GHCIDR uses sampled images from the full dataset. Aggregation of images causes encapsulation of features, which makes it difficult for the model to learn the important features. The accuracy of GHCIDR is roughly equal to that of the full dataset. Thus, we can replace the full-dataset with the reduced set without compromising accuracy. By tuning the parameters - $\alpha$ \& $\beta$, we can achieve more accuracy but at the cost of a lesser reduction rate. 

The training time of datasets generated from Merged-GHCIDR was comparable to that of RHC and significantly lesser than that required on the full dataset. We improved the training time of MNIST on VGG1 to about ${1/7}^{th}$ of the time required on the full dataset. Similarly, the training time of FMNIST on FCN and CIFAR10 on VGG1 were reduced to around ${1/3}^{rd}$ and ${2/3}^{rd}$ of the time for the full dataset, respectively.